\pdfoutput=1
\RequirePackage{fix-cm}%
\documentclass[10pt,twocolumn,letterpaper]{article}
\ifdefined\ETReview
\usepackage[review]{cvpr}
\providecommand{\confName}{CVPR}
\providecommand{\confYear}{2027}
\else
\usepackage[pagenumbers]{cvpr}
\fi
\usepackage[T1]{fontenc}
\usepackage{amsmath,amssymb,bm,mathtools}
\usepackage[table]{xcolor}
\usepackage{graphicx,booktabs,array,multirow,microtype,pifont}
\usepackage[noend]{algpseudocode}
\usepackage{algorithm}
\algrenewcommand\algorithmicrequire{\textbf{Input:}}
\algrenewcommand\algorithmicensure{\textbf{Output:}}
\usepackage{enumitem}
\usepackage{placeins}
\usepackage{tikz}
\usepackage{pgfplots}
\pgfplotsset{compat=1.17}
\usetikzlibrary{arrows.meta,positioning,calc,fit,backgrounds,decorations.pathreplacing,shapes.geometric}
\definecolor{etBlue}{HTML}{08306B}   %
\definecolor{etOrange}{HTML}{FF851B} %
\definecolor{etAqua}{HTML}{3787C0}   %
\definecolor{etYellow}{HTML}{EDA100}
\definecolor{etMagenta}{HTML}{E87BA4}
\definecolor{etGreen}{HTML}{008300}
\definecolor{etNavy}{HTML}{184F95}
\definecolor{etLight}{HTML}{EAF2FC}
\definecolor{etInk}{HTML}{24313F}
\definecolor{etGray}{HTML}{77818E}
\definecolor{etPale}{HTML}{F2F3F5}
\definecolor{etSky}{HTML}{DAE8FB}   %
\definecolor{etPeach}{HTML}{FFDDB5} %
\newcommand{\best}[1]{\textbf{#1}}
\newcommand{\second}[1]{\underline{#1}}
\newcommand{\oursmark}{~(ours)}
\newlength{\ETwhiterulewidth}
\newcommand{\whiterule}{\arrayrulecolor{white}\specialrule{\ETwhiterulewidth}{0pt}{0pt}\arrayrulecolor{black}}
\definecolor{lolGroup}{HTML}{EFEFEF} %
\definecolor{tabblue}{HTML}{5555CC} %
\definecolor{lolNavy}{HTML}{08306B}\definecolor{lolMid}{HTML}{3787C0}\definecolor{lolLight}{HTML}{ABD0E6}
\definecolor{lolB1}{HTML}{C6DBEF}\definecolor{lolB2}{HTML}{6BAED6}\definecolor{lolB3}{HTML}{2171B5}
\definecolor{lolOrange}{HTML}{FF851B}\definecolor{lolRed}{HTML}{D83233}
\tikzset{
 etbox/.style={draw=etGray,line width=.55pt,rounded corners=2pt,align=center,fill=white,font=\fontsize{7}{8.5}\selectfont,text=etInk},
 etours/.style={etbox,draw=etBlue,fill=etLight},
 etsm/.style={etbox,fill=etPale,minimum width=1.25cm,minimum height=.30cm,inner sep=1pt},
 etrec/.style={etours,minimum width=1.25cm,minimum height=.30cm,inner sep=1pt},
 etflow/.style={-{Stealth[length=3.2pt]},draw=etInk,line width=.6pt},
 etmatch/.style={{Stealth[length=3pt]}-{Stealth[length=3pt]},draw=etGray,dashed,line width=.6pt},
 ettitle/.style={font=\bfseries\fontsize{8}{9.5}\selectfont,text=etInk},
 etnote/.style={font=\itshape\fontsize{6.5}{8}\selectfont,text=etGray,align=center},
 alien/.pic={
  \draw[etNavy,line width=.7pt] (-.12,.25)--(-.25,.46) (.12,.25)--(.25,.46);
  \fill[etBlue] (-.25,.46) circle(.045) (.25,.46) circle(.045);
  \path[fill=etBlue,draw=etNavy,line width=.6pt]
   (0,.30) .. controls (.38,.30) and (.39,-.03) .. (.18,-.22)
   .. controls (.08,-.31) and (-.08,-.31) .. (-.18,-.22)
   .. controls (-.39,-.03) and (-.38,.30) .. (0,.30)--cycle;
  \fill[white,rotate around={-28:(.13,.04)}] (.13,.04) ellipse(.105 and .06);
  \fill[white,rotate around={28:(-.13,.04)}] (-.13,.04) ellipse(.105 and .06);
  \draw[etNavy,line width=.65pt] (-.06,-.16)..controls(0,-.19)..(.06,-.16);
 },
}

\pgfplotsset{
 etaxis/.style={
  axis line style={black!75,line width=.45pt}, tick style={black!60},
  ticklabel style={font=\tiny,text=etInk}, label style={font=\scriptsize,text=etInk},
  title style={font=\scriptsize\bfseries,text=etInk,yshift=-4pt},
  grid=major, grid style={black!13,line width=.35pt},
  legend style={font=\tiny,draw=black!22,rounded corners=1pt,fill=white,fill opacity=.9,text opacity=1,row sep=-2.5pt,inner sep=1.5pt},
  legend cell align=left, every axis plot/.append style={line width=.9pt},
 },
 ours/.style={etBlue,mark=*,mark size=1.4pt,line width=1.3pt},
 tttc/.style={lolOrange,mark=triangle*,mark size=1.6pt},
 glac/.style={etAqua,mark=square*,mark size=1.1pt},
 mambatwoc/.style={etYellow,mark=diamond*,mark size=1.5pt},
 mambathreec/.style={etMagenta,mark=pentagon*,mark size=1.4pt},
 gdnoffc/.style={lolRed,mark=x,mark size=1.6pt},
 etbarlegend/.style={legend image code/.code={\draw[##1] (0cm,-0.08cm) rectangle (0.22cm,0.08cm);}},
 teacherc/.style={lolLight,dashed,line width=1pt,mark=*,mark size=1.1pt,mark options={solid}},
}

\newcommand{\fitto}[2]{\resizebox{\ifdim\width>#1 #1\else\width\fi}{!}{#2}}
\makeatletter
\def\ET@rowfill{{\let\CT@color\color\CT@row@color\vrule\@width.5\ETwhiterulewidth}}
\protected\def\ET@vrule{\ifx\CT@row@color\relax\CT@arc@\vline\else
  \kern-.5\ETwhiterulewidth\ET@rowfill{\CT@arc@\vline}\ET@rowfill\kern-.5\ETwhiterulewidth\fi}
\def\@arrayrule{\@addtopreamble{{\ET@vrule}}}
\def\CT@rowc{\@ifnextchar[\CT@rowd{\CT@rowe[\dimexpr\col@sep-.5\ETwhiterulewidth\relax][\dimexpr\col@sep-.5\ETwhiterulewidth\relax]}}
\makeatother
\usetikzlibrary{arrows.meta,calc}
\definecolor{lolBlue}{HTML}{0000FF}
\definecolor{lolCol}{HTML}{CFE2F3}
\definecolor{lolColGray}{HTML}{EFEFEF}
\definecolor{lolQuote}{HTML}{666666}
\definecolor{lolQuoteLight}{HTML}{B7B7B7}
\definecolor{lolSnow}{HTML}{4295FF}
\definecolor{lolLoss}{HTML}{660000}
\definecolor{lolDeep}{HTML}{20124D}
\newcommand{\lolf}[2]{\fontencoding{T1}\fontfamily{Roboto-LF}\fontseries{#1}\fontshape{n}\fontsize{#2}{#2}\selectfont}
\newcommand{\lolfi}[2]{\fontencoding{T1}\fontfamily{Roboto-LF}\fontseries{#1}\fontshape{it}\fontsize{#2}{#2}\selectfont}

\definecolor{lolLabelBg}{HTML}{F4F6FB}
\tikzset{
 lol/.style={x=1pt,y=-1pt,line cap=butt,line join=miter},
 lolframe/.style={draw=black,line width=2.25pt,fill=white},
 loltitle/.style={anchor=base,inner sep=0pt,font=\lolf{b}{56}},
 lolsub/.style={anchor=base,inner sep=0pt,font=\lolf{m}{42}},
 lollab/.style={anchor=base,inner sep=0pt,font=\lolf{m}{36}},
 lolquote/.style={anchor=base,inner sep=0pt,font=\lolfi{m}{36},text=lolQuote},
 lolsmall/.style={anchor=base,inner sep=0pt,font=\lolf{l}{19}},
 lolpass/.style={anchor=base,inner sep=0pt,font=\lolf{l}{20}},
 lolxn/.style={anchor=base,inner sep=0pt,font=\lolf{l}{48}},
 lolmath/.style={inner sep=0pt,font=\lolfi{l}{42}},
 lolmathS/.style={inner sep=0pt,font=\lolfi{l}{36}},
 lolline/.style={line width=2.25pt},
 lolwire/.style={line width=3pt},
 lolarr/.style={-{Triangle[width={7.4pt 0},length={10.2pt 0}]},line width=2.25pt},
 lolarrB/.style={-{Triangle[width={9.9pt 0},length={13.6pt 0}]},line width=3pt},
 loltrap/.style={draw=black,line width=3pt,fill=white,line join=miter},
}

\newcommand{\loltrap}[7][]{%
 \if#7d\draw[loltrap,#1] (#2,#3)--(#4,#3)--({#4-(#4-#2)*#6},#5)--({#2+(#4-#2)*#6},#5)--cycle;%
 \else\draw[loltrap,#1] (#2,#5)--({#2+(#4-#2)*#6},#3)--({#4-(#4-#2)*#6},#3)--(#4,#5)--cycle;\fi}
\definecolor{etfFit}{HTML}{EB6834}%
\tikzset{
 etnew/.style={preaction={draw=etfFit,opacity=.30,line width=12pt}},
 snew/.style={etnew}, sscan/.style={etnew}, fnew/.style={etnew}, fscan/.style={etnew}, pnew/.style={etnew}, pscan/.style={etnew},
}

\usepackage[breaklinks,colorlinks,linkcolor=blue,citecolor=tabblue,urlcolor=blue]{hyperref}
\hypersetup{pdftitle={ET: Efficient Transfer from Softmax to Hybrid Attention},pdfauthor={Guangfu Guo, Xiaoqian Lu, Linsey Pang, Weiran Yao, Haolin Chen, Long Cheng, Kunpeng Liu}}
\makeatletter
\AtBeginDocument{\def\theHALG@line{\thealgorithm.\arabic{ALG@line}}}
\makeatother
\newcommand{\ours}{ET}               %
\newcommand{\oursplain}{ET}

\newcommand{\etpG}{ET$^{+}_{\mathrm{G}}$}

\newcommand{\etpD}{ET$^{+}_{\mathrm{D}}$}

\newcommand{\tfive}{T$^5$}
\newcommand{\tfivep}{T$^{5+}$}

\makeatletter\newcommand\captionof[1]{\def\@captype{#1}\caption}\makeatother %

\AtBeginDocument{\crefname{algorithm}{Alg.}{Algs.}\Crefname{algorithm}{Algorithm}{Algorithms}}
\title{ET: Efficient Transfer from Softmax to Hybrid Attention}
\author{\begin{tabular}{c}
Guangfu Guo\textsuperscript{1,5,$\dagger$}, Xiaoqian Lu\textsuperscript{2,$\dagger$}, Linsey Pang\textsuperscript{3},
Weiran Yao\textsuperscript{4},\\[2pt]
Haolin Chen\textsuperscript{4}, Long Cheng\textsuperscript{1,*}, Kunpeng Liu\textsuperscript{1,5,*}\\[3pt]
\textsuperscript{1}Clemson University\quad
\textsuperscript{2}University of Ottawa\quad
\textsuperscript{3}PayPal\\[2pt]
\textsuperscript{4}actAVA AI\quad
\textsuperscript{5}CLAIM Lab\\[2pt]
{\small\texttt{gguo@clemson.edu}, \texttt{xlu053@uottawa.ca}, \texttt{panglinsey@gmail.com}}\\
{\small\texttt{weiran@actava.ai}, \texttt{haolin@actava.ai},
\texttt{lcheng2@clemson.edu}, \texttt{kunpenl@clemson.edu}}\\[-1pt]
{\small\textsuperscript{$\dagger$}Equal contribution\quad\textsuperscript{*}Corresponding authors}
\end{tabular}}
\begin{document}
\maketitle
\begin{abstract}
\noindent Converting pretrained softmax Transformers to efficient attention often causes substantial performance degradation, while recovery typically requires costly fine-tuning.
We propose \textbf{ET (Efficient Transfer)}, a framework for high-fidelity attention conversion with minimal optimization cost.
ET combines exact local and shared-key softmax attention with a mean-anchored, scan-free global linear correction and adaptive gating, achieving linear complexity in sequence length.
To preserve pretrained knowledge, ET freezes pretrained weights and calibrates only newly introduced parameters through alternating least squares on a small set of unlabeled samples, without gradient updates.
On ImageNet-1K, ET achieves 79.64\% and 81.55\% top-1 accuracy on DeiT-S and DeiT-B, respectively, within 0.25 percentage points of their softmax teachers.
At $768{\times}768$, ET outperforms \tfive{} by 3.59 percentage points at comparable inference FLOPs, without gradient-based fine-tuning.
On DiT-XL/2, ET achieves an FID-50K of 2.59 with only 250 LoRA updates.
On Stable Diffusion 3.5 Medium, ET matches the teacher's GenEval score at 1K resolution with just 0.73 A100 GPU-hours of gradient-free conversion and a 1.39$\times$ inference speedup.
\end{abstract}
\section{Introduction}
\label{sec:intro}
In recent years, Vision Transformers (ViTs)~\cite{vit,deit} have achieved remarkable success in image classification, visual representation learning, and image generation~\cite{guo2026visualrepresentationhistorymodeling}.
However, the quadratic complexity of softmax attention with respect to sequence length leads to substantial computational costs, particularly at high resolutions.
Linear attention and other efficient attention mechanisms~\cite{performer,linear_transformers,gla,gdn} offer promising alternatives for reducing inference costs.
Nevertheless, directly replacing softmax attention in pretrained Transformers often causes severe performance degradation, while training efficient architectures from scratch requires substantial computational resources.
Efficiently transferring pretrained knowledge to alternative attention architectures, while preserving model performance and minimizing conversion costs, therefore remains an important challenge.

Several studies have explored converting pretrained softmax Transformers into efficient attention models.
For example, LiT~\cite{lit} leverages pretrained weight initialization and knowledge distillation to convert Diffusion Transformers (DiTs) into linear attention models; \tfive{}~\cite{t5paper} introduces a Test-Time Training (TTT) architecture and recovers performance through fine-tuning; and LoLCATs~\cite{lolcats} performs attention conversion through attention output matching and low-rank adaptation (LoRA)~\cite{lora}.
These approaches demonstrate the effectiveness of leveraging pretrained knowledge to reduce the cost of developing efficient attention models.
However, recovering performance after architectural conversion often still requires substantial gradient-based optimization.
Meanwhile, gradient-free conversion methods~\cite{monarchattention} avoid expensive fine-tuning but can suffer considerable accuracy degradation when replacing all attention layers.
These limitations highlight the need for an attention transfer framework that preserves pretrained capabilities with minimal optimization overhead while providing a favorable accuracy--efficiency trade-off.

\begin{figure*}[t]
\centering
\begin{minipage}[c]{.398\linewidth}
\centering
\includegraphics[width=\linewidth]{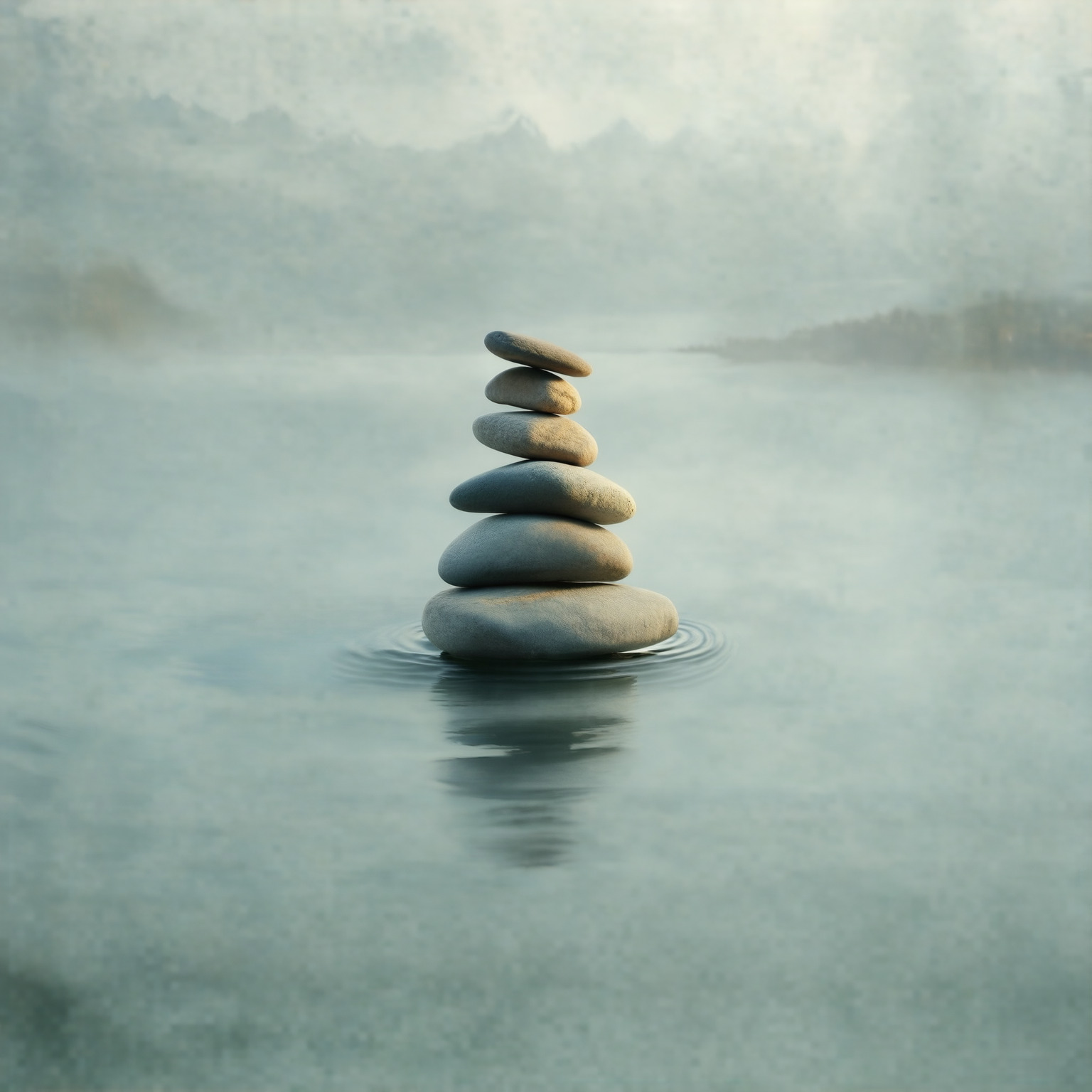}
\end{minipage}\hfill
\begin{minipage}[c]{.598\linewidth}
\centering
\includegraphics[width=.33\linewidth]{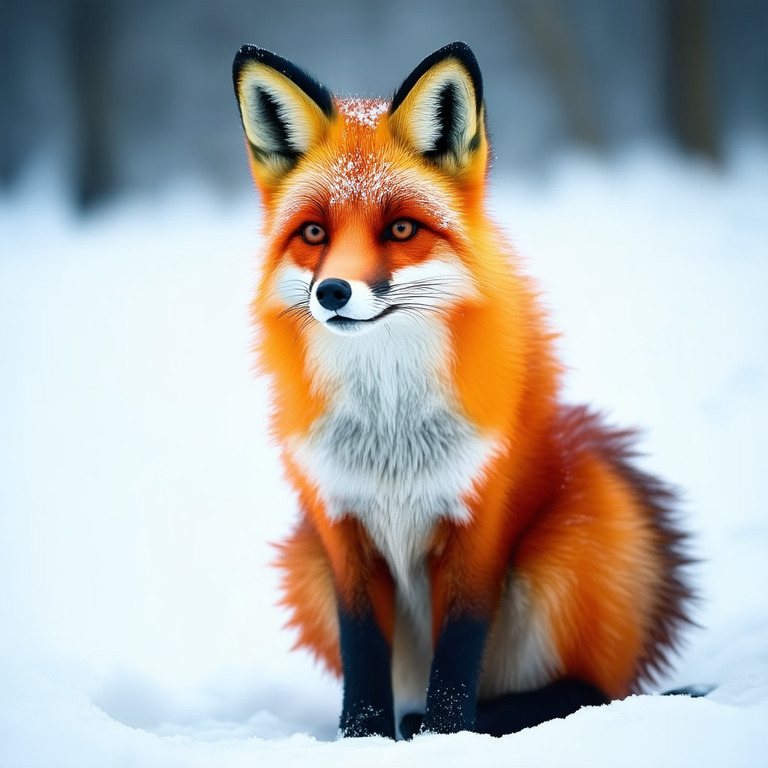}\hfill
\includegraphics[width=.33\linewidth]{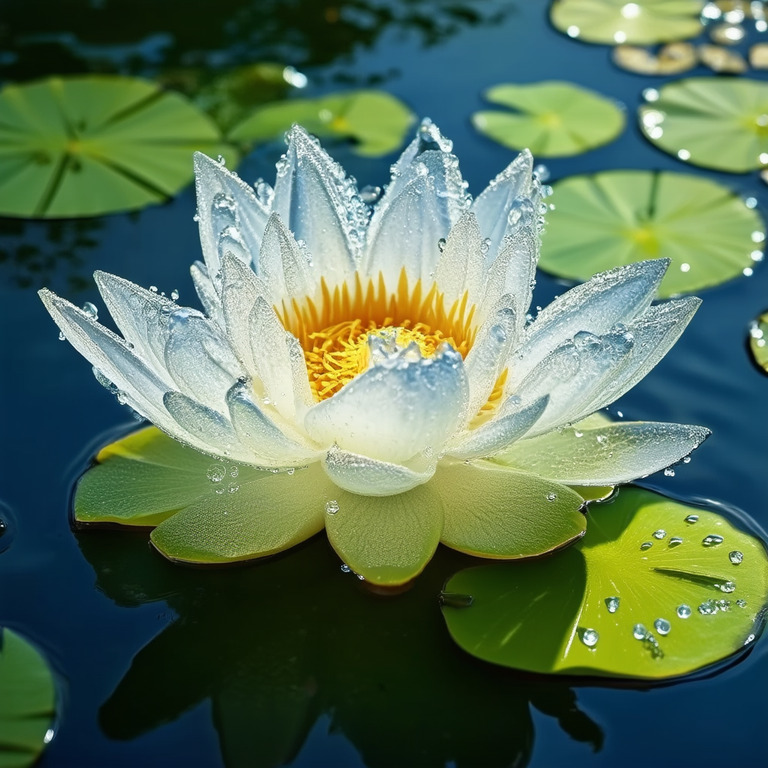}\hfill
\includegraphics[width=.33\linewidth]{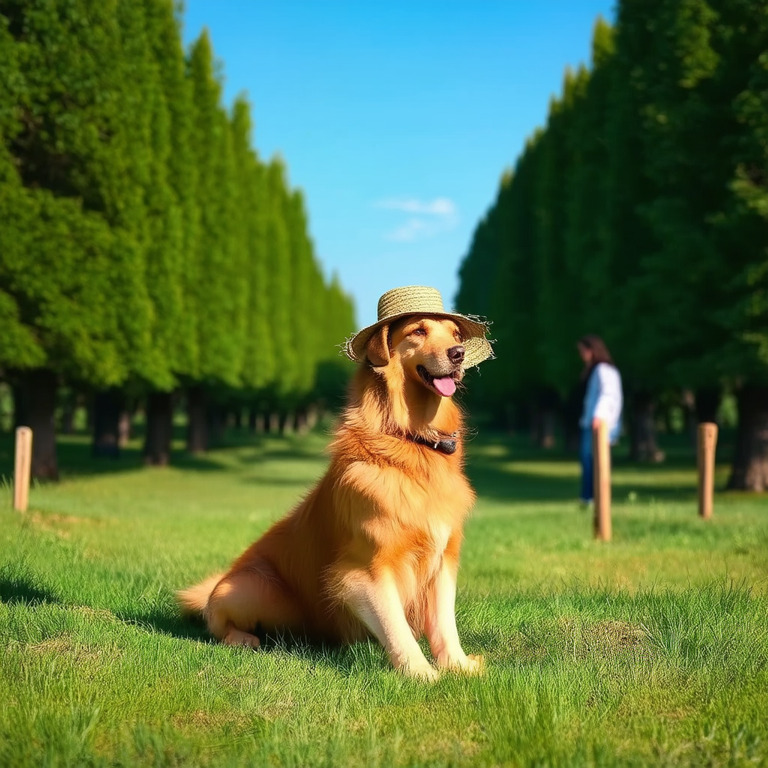}\par\vspace{1pt}
\includegraphics[width=.33\linewidth]{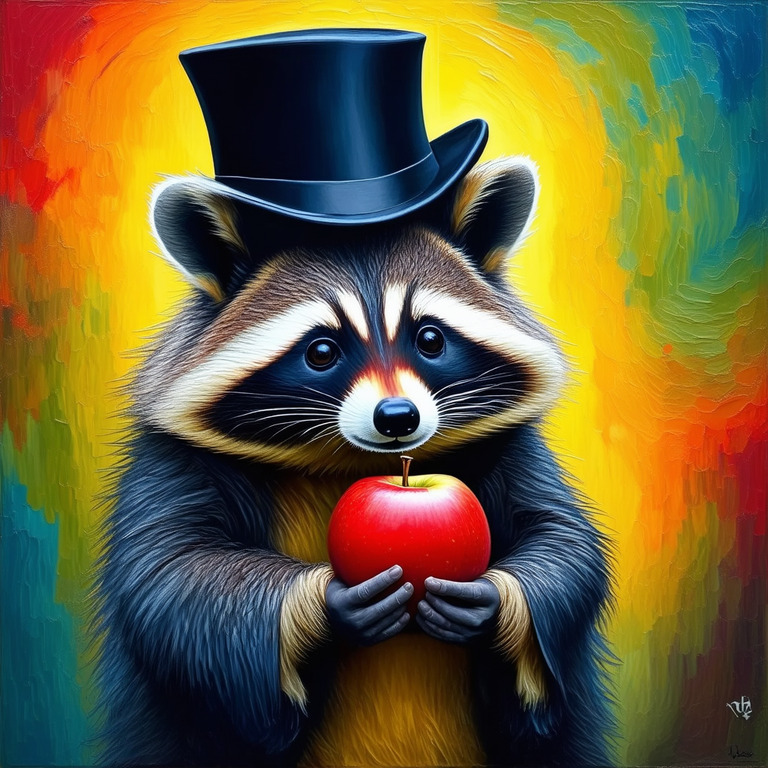}\hfill
\includegraphics[width=.33\linewidth]{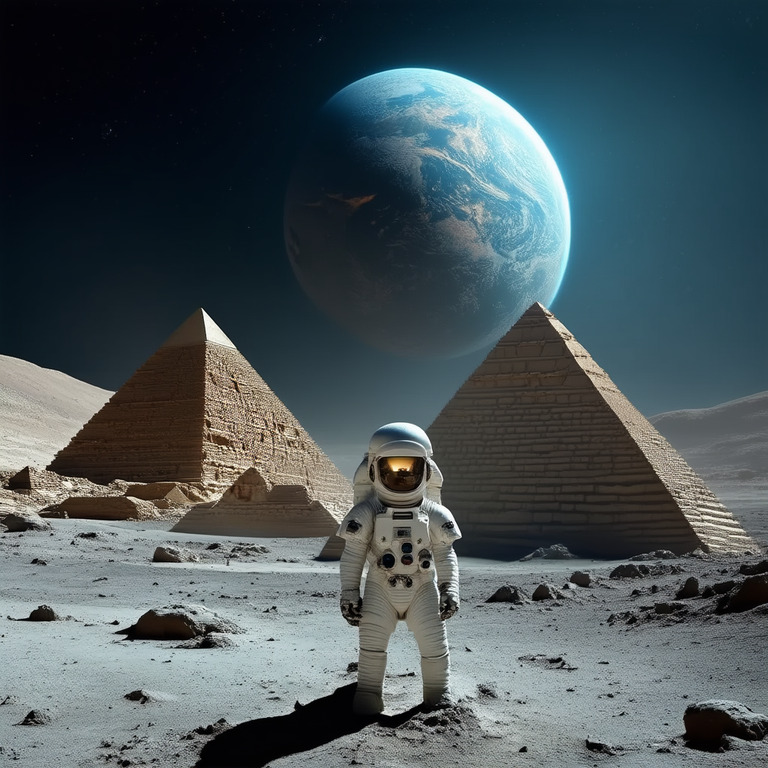}\hfill
\includegraphics[width=.33\linewidth]{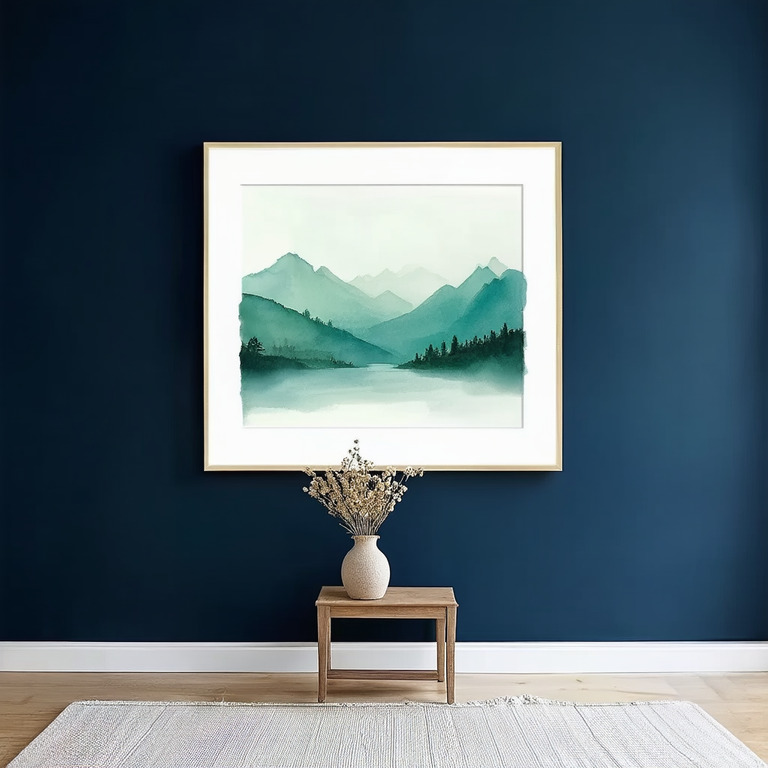}
\end{minipage}
\caption{Left: $2048^2$ image generated by SD3.5-\etpD{} (ET with a Delta-rule scan); right: $1024^2$ images generated by SD3.5-\etpD{}. \etpD{} is converted without gradients and then adapted with 800 LoRA updates.}
\label{fig:sd35_showcase}
\end{figure*}

In this paper, we investigate how to efficiently transfer pretrained softmax Transformers to hybrid attention architectures while minimizing performance degradation and optimization costs.
Specifically, we address two key questions: \textbf{(1) How can we design an efficient hybrid attention mechanism that preserves the essential interactions of the original softmax attention? (2) How can we leverage pretrained knowledge to recover model performance with minimal calibration and fine-tuning costs?}
To this end, we propose \textbf{ET} (Efficient Transfer), a framework that combines high-fidelity hybrid attention design with lightweight, gradient-free knowledge transfer.

\begin{figure*}[t]
\centering
\includegraphics[width=\textwidth]{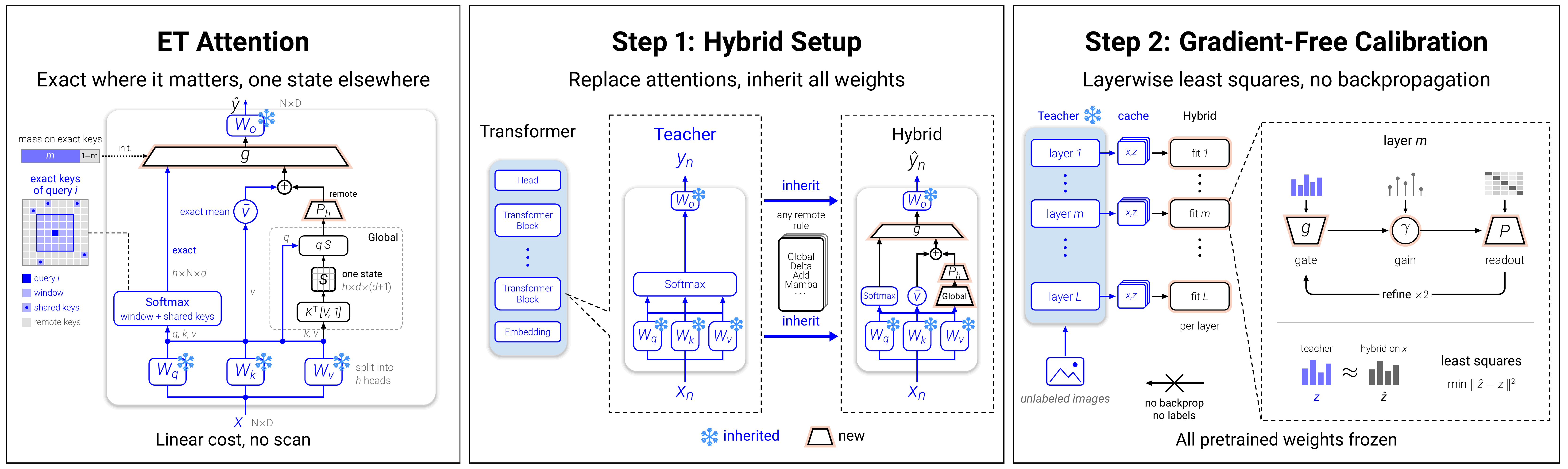}
\vspace{-0.45cm}
\caption{Overview of \ours{}. Each softmax attention becomes a hybrid attention (left): exact attention on a local window and a few shared keys, plus the exact remote mean and a correction read from one shared state $\bm S$, mixed by a gate. Step~1 constructs the hybrid architecture while inheriting pretrained weights. Step~2 calibrates the newly introduced parameters through layer-specific least squares without backpropagation.}
\label{fig:pull_method}
\vspace{-0.25cm}
\end{figure*}

Our core idea is to decompose attention into two complementary components: exact computation and efficient approximation (\cref{fig:pull_method}).
Specifically, ET preserves exact softmax attention within local windows and over a small set of shared keys to retain important local and global interactions.
For the remaining positions, ET preserves the exact mean of remote values and approximates only the query-dependent deviation through a scan-free global linear correction.
Adaptive gating combines the exact and approximate branches, enabling efficient attention computation with linear complexity in sequence length.
To transfer pretrained knowledge efficiently, ET inherits and freezes all original model weights and calibrates only newly introduced parameters through alternating least squares using a small number of unlabeled samples, without backpropagation or gradient updates.
For challenging generative tasks, where approximation errors may accumulate across layers and sampling steps, the converted model further provides a high-quality initialization for lightweight LoRA recovery.
ET thus integrates high-fidelity attention approximation and low-cost knowledge transfer within a unified framework.

We systematically evaluate ET on image classification, class-conditional image generation, and text-to-image generation.
On ImageNet-1K, ET achieves 79.64\% and 81.55\% top-1 accuracy on DeiT-S and DeiT-B, respectively, after converting all attention layers without gradient updates, with only 0.19 and 0.25 percentage point drops from their softmax teachers (\cref{tab:cls_clsmain}).
Calibration of DeiT-S takes only 31 seconds on a single A100 GPU (\cref{sec:cost}).
At $768{\times}768$, ET outperforms \tfive{} by 3.59 percentage points at comparable inference floating-point operations (FLOPs), while requiring 29\% fewer multiply--accumulate operations (MACs) than the softmax teacher (\cref{fig:budget}).
On DiT-XL/2~\cite{dit}, only 250 LoRA updates reduce the Fr\'echet Inception Distance on 50K samples (FID-50K) from 5.61 to 2.59, approaching the 2.35 achieved by the softmax teacher under the same recovery budget (\cref{tab:t5dit}).
On Stable Diffusion 3.5 Medium (SD3.5-Medium)~\cite{sd3}, ET achieves a GenEval score of 0.666 at 1K resolution without gradient updates, matching the teacher's 0.665, with only 0.73 A100 GPU-hours of conversion cost.
Meanwhile, ET accelerates inference by $1.39\times$ and $1.68\times$ at 1K and 2K resolutions, respectively (\cref{tab:t5sd35}).

Our main contributions are summarized as follows:
\begin{enumerate}[leftmargin=*,itemsep=0pt,topsep=1pt]
\item We propose \textbf{ET}, a high-fidelity hybrid attention framework that combines exact local and shared-key softmax attention with mean-anchored remote approximation and scan-free global correction, enabling linear complexity in sequence length.
\item We develop an efficient \textbf{gradient-free calibration strategy} that transfers pretrained attention through alternating least squares on a small number of unlabeled samples, without updating pretrained weights, while supporting further performance recovery through lightweight LoRA adaptation.
\item Extensive experiments across classification and generative tasks demonstrate that ET achieves high-fidelity attention conversion with minimal optimization cost, favorable accuracy--efficiency trade-offs under constrained inference computation, and practical inference speedups at high resolutions.
\end{enumerate}
\section{Related Work}
\label{sec:related}
\paragraph{Efficient and hybrid attention.}
To reduce the cost of softmax attention, prior work has explored kernelized linear attention~\cite{performer,linear_transformers}, low-rank approximation~\cite{nystromformer} and learnable feature maps~\cite{hedgehog}.
Neighborhood Attention~\cite{natten} strengthens spatial modeling with local attention, while Based~\cite{based}, STILL~\cite{still} and Sparse--Linear Attention (SLA)~\cite{sla} further combine exact attention with a linear branch.
Dynamic Linear Attention (DLA)~\cite{dynamiclinear} increases expressiveness through dynamic state fusion, and recurrent architectures such as Gated Linear Attention (GLA)~\cite{gla}, Mamba-2~\cite{mamba2}, DeltaNet~\cite{deltanet} and Gated DeltaNet~\cite{gdn} and their bidirectional extensions~\cite{lion,hydra,vim,deltaflow} model sequences efficiently.

\paragraph{Attention conversion of pretrained Transformers.}
LiT~\cite{lit} converts diffusion Transformers into linear-attention models through weight inheritance and knowledge distillation; \tfive{}~\cite{t5paper} converts vision Transformers with test-time training and fine-tuning; and ViT-AdaLA~\cite{vitadala} assists conversion through attention and feature alignment.
In language models, LoLCATs~\cite{lolcats}, Liger~\cite{liger} and MOHAWK~\cite{mohawk} adopt attention-output matching, gate initialization and staged distillation, respectively.
CLEAR~\cite{clear}, ReHyAt~\cite{rehyat}, Attention Surgery~\cite{attention_surgery} and LinVideo~\cite{linvideo} extend efficient-attention conversion to image and video generation.
These methods usually require distillation or fine-tuning to recover the performance of the converted model.

\paragraph{Low-cost and analytic conversion.}
MonarchAttention (MA)~\cite{monarchattention} replaces attention without additional training through an input-dependent structured approximation, and Fast Nystr\"om Attention (FNA)~\cite{lu2025artifacts} explores structured attention approximation for vision Transformers.
Neural Block Linearization (NBL)~\cite{nbl} replaces selected attention layers with a closed-form linear minimum-mean-square-error estimator, while Taylor-Calibrate~\cite{taylorcalibrate} initializes hybrid attention models from attention statistics and then aligns them with gradient-based optimization.
\ours{} combines exact remote-mean preservation with a shared scan-free key--value state and closed-form calibration of the newly introduced learnable parameters.
\section{Method}
\label{sec:method}\label{sec:preliminaries}
\ours{} (Efficient Transfer) has two components (\cref{fig:pull_method}): \emph{mean-anchored hybrid attention} (\cref{sec:adapt}), which keeps important local and shared-key interactions exact and represents remote attention as an exact value mean plus a calibrated global correction, and \emph{gradient-free calibration} (\cref{sec:method_framework}), which keeps all pretrained weights and fits only the new parameters by alternating least squares; optional LoRA recovers harder generative tasks.

\subsection{Mean-Anchored Hybrid Attention}
\label{sec:adapt}\label{sec:method_new_stuff}
\paragraph{Exact attention decomposition.}
Consider one attention head of width $d$, with softmax attention $\bm y_i=\sum_j a_{ij}\bm v_j$, where $a_{ij}=\exp(\bm q_i^\top\bm k_j/\sqrt d)/\sum_l\exp(\bm q_i^\top\bm k_l/\sqrt d)$.
For each patch query $i$, we partition the keys into two disjoint sets: an exact set $\mathcal E(i)$ and a remote set $\mathcal R(i)$.
The exact set contains a $k{\times}k$ local window following Neighborhood Attention (NAT)~\cite{natten}, $s$ shared keys selected per image and head according to the largest $\bm k_j^\top\bar{\bm q}$, and special tokens such as the class token when present; the remaining keys form $\mathcal R(i)$.
We denote this configuration by NAT$k$+$s$ (NAT$k$ without shared keys). Our default NAT7+32 uses 81 nominal key slots, about 41\% of the 196 image patches at $224{\times}224$ before accounting for overlaps.
Let $Z_i^{\mathcal E}$ and $Z_i^{\mathcal R}$ denote the softmax partition functions over the two sets, and $\bm y_i^{\mathcal E}$ and $\bm y_i^{\mathcal R}$ their separately normalized attention outputs. Full softmax attention admits the exact decomposition
\begin{equation}
\boxed{\bm y_i=m_i\,\bm y_i^{\mathcal E}+(1-m_i)\,\bm y_i^{\mathcal R},\qquad
m_i=\frac{Z_i^{\mathcal E}}{Z_i^{\mathcal E}+Z_i^{\mathcal R}}.}
\label{eq:decomp}
\end{equation}

\noindent\textbf{Proposition 1 (exact attention decomposition).}
\textit{\Cref{eq:decomp} holds for any non-empty, disjoint partition $\mathcal E(i)$, $\mathcal R(i)$} (split the softmax numerator and normalizer over the two sets; full proofs of all propositions are in the supplementary material).
\ours{} therefore preserves $\bm y_i^{\mathcal E}$ exactly and approximates only $\bm y_i^{\mathcal R}$ and $m_i$.

\paragraph{Mean-anchored remote approximation.}
For each query $i$, let $\bar{\bm v}_{\mathcal R}(i)$ denote the mean value of the remote set $\mathcal R(i)$. We compute this mean exactly from the total value sum by subtracting the contributions of the exact set $\mathcal E(i)$, and represent the remote attention output as this exact mean plus a calibrated residual, $\hat{\bm y}_i^{\mathcal R}=\bar{\bm v}_{\mathcal R}(i)+\gamma_h\bm P_h\bm c_i$.

\paragraph{Scan-free global correction.}
To construct the residual features efficiently, \ours{} aggregates all key--value pairs into a single global state shared across queries:
\begin{equation}
\bm S_h=\sum_{j=1}^{N}\tilde{\bm k}_j\,[\bm v_j;1]^\top\in\mathbb R^{d\times(d+1)},\qquad
[\bm o_i;o_i^{(1)}]=\frac1N\bm S_h^\top\hat{\bm q}_i,
\label{eq:state}
\end{equation}
where $\tilde{\bm k}_j$ is centered by the per-image key mean and $\ell_2$-normalized, $\hat{\bm q}_i=s_h\bm q_i/\sqrt d$, and the per-head scale $s_h$ is the root-mean-square norm of the centered keys of head $h$ on the calibration data. For each query, the state produces a value readout $\bm o_i$ and a normalization readout $o_i^{(1)}$, which are combined with the exact remote mean to form the centered correction
\begin{equation}
\boxed{\bm c_i=\bm o_i-o_i^{(1)}\,\bar{\bm v}_{\mathcal R}(i).}
\label{eq:far}
\end{equation}
The global state captures key--value interactions across the image, while the query-dependent remote mean provides the reference for residual reconstruction. The gain $\gamma_h$ and readout $\bm P_h$ are calibrated to approximate the remote attention deviation using cached teacher outputs. The resulting hybrid attention is
\begin{equation}
\boxed{\hat{\bm y}_i^{\mathcal R}=\bar{\bm v}_{\mathcal R}(i)+\gamma_h\bm P_h\bm c_i,\ \
\hat{\bm y}_i=g_i\bm y_i^{\mathcal E}+(1-g_i)\hat{\bm y}_i^{\mathcal R},}
\label{eq:hybrid}
\end{equation}
where $g_i$ is an adaptive gate.

\noindent\textbf{Proposition 2 (constant preservation and translation equivariance).}
\textit{With queries, keys, gates and fitted parameters fixed, \ours{} reproduces any constant value input and is equivariant to a uniform translation of all values.}

\textit{Proof sketch.} Shifting all values by $\bm a$ adds $o_i^{(1)}\bm a$ to $\bm o_i$ and $\bm a$ to $\bar{\bm v}_{\mathcal R}(i)$, so $\bm c_i$ is unchanged; the mixing weights sum to one.

\paragraph{Adaptive gating.}
By \cref{eq:decomp}, the ideal gate is the softmax mass of the exact set. \ours{} computes $Z_i^{\mathcal E}$ exactly and approximates $Z_i^{\mathcal R}$ by a cheap positive proxy $\phi_i$ (remote count, mean key or exponential linear unit (ELU)$+1$ feature) with a calibrated residual:
\begin{equation}
\boxed{g_i=\sigma\!\left(\log Z_i^{\mathcal E}-\log\phi_i-\bm w_h^\top\bm x_i-b_h\right),}
\label{eq:massgate}
\end{equation}
where $\bm x_i$ is the input token before the frozen query/key/value (QKV) projections and $\bm w_h,b_h$ are new gate parameters.
With target $t_i=\log Z_i^{\mathcal R}-\log\phi_i$, the $\frac14$-Lipschitz sigmoid gives $|g_i-m_i|\le\frac14|t_i-\bm w_h^\top\bm x_i-b_h|$, so fitting $t_i$ initializes the gate; calibration then refines it for output reconstruction.

\paragraph{Computational complexity.}
For fixed $k$, $s$, $d$, each head costs $O(N(k^2{+}s)d+Nd^2)$, linear in $N$: the exact branch $O(N(k^2{+}s)d)$, the shared state $O(Nd^2)$; the window terms of the remote mean use integral-image box sums, and one shared state needs no recurrent scan.

\subsection{Efficient Calibration and Recovery}
\label{sec:method_framework}\label{sec:refine}
\begin{algorithm}[t]
\caption{Gradient-free conversion with \ours{}. Lines 1--11 perform gradient-free calibration; line 12 performs optional gradient-based recovery.}
\label{alg:et}
\small
\begin{algorithmic}[1]
\Require teacher with $M$ attention layers; unlabeled images $\mathcal D$
\Require exact set NAT$k{+}s$; refinement rounds $R{=}2$
\Ensure \ours{} model; all inherited weights unchanged
\State run teacher on $\mathcal D$; cache $(\bm x^m_T,\bm z^m_T)$, all $m$ \Comment{no grad}
\For{$m=1,\dots,M$}
  \State $\bm x^m\gets$ DiT: input of the partially converted model; else $\bm x^m_T$; target $\bm z^m_T$
  \State build $\mathcal E(i)$, $\bar{\bm v}_{\mathcal R(i)}$, $\bm S$; set $s_h$ \Comment{\cref{eq:decomp,eq:state}}
  \State per $\phi$: ridge $(\bm w_h,b_h)$, $g_i\approx m_i$; keep best \Comment{\cref{eq:massgate}}
  \State $\gamma_h\gets$ weighted least squares to $\bm y^{\mathcal R}-\bar{\bm v}_{\mathcal R}$
  \State $\bm P_h\gets$ ridge to $\bm z^m_T$ through $\bm W_o^m$
  \For{$r\in[R]$}
    \State $u_i\gets$ $H{\times}H$ ridge on the projected output; $g_i^\star\gets\operatorname{clip}(u_i,\varepsilon,1{-}\varepsilon)$
    \State refit $(\bm w_h,b_h)$ to $\operatorname{logit}g_i^\star$, then $\bm P_h$
    \State keep the round only if $\mathcal L_m$ drops \Comment{\cref{eq:relmse}}
  \EndFor
\EndFor
\State \textit{optional:} fine-tune LoRA $\cup$ new \Comment{uses gradients}
\end{algorithmic}
\end{algorithm}
\paragraph{Gradient-free least-squares calibration.}
On a small unlabeled set $\mathcal D$, the teacher's cached projected attention outputs $\bm Z_{\mathrm P}^m$ are the targets of layer $m$:
\begin{equation}
\boxed{\mathcal L_m(\theta_m)=\frac{\sum_{x\in\mathcal D}\|\bm Z_{\mathrm P}^m(x)-\hat{\bm Z}_{\mathrm P}^m(x;\theta_m)\|_F^2}{\sum_{x\in\mathcal D}\|\bm Z_{\mathrm P}^m(x)\|_F^2},}
\label{eq:relmse}
\end{equation}
where $\theta_m$ holds the new gate, gain and readout parameters; pretrained weights, including the output projection, stay frozen.

\paragraph{Layerwise calibration.}
For classification and SD3.5, each attention layer is calibrated independently using cached teacher inputs and projected attention outputs. For DiT, we adopt sequential layerwise calibration: after converting the preceding layers, we take the input of the current layer from the partially converted model and use the corresponding cached teacher attention output as the regression target (\cref{sec:gen}). This exposes each layer's calibration to the discrepancies introduced by the preceding converted layers and aligns its output with the original teacher trajectory. All steps operate on cached or forward-computed activations without backpropagation through the network.

\paragraph{Closed-form parameter calibration.}
\ours{} freezes all pretrained projections and calibrates the newly introduced gate parameters $(\bm w_h,b_h)$, correction gain $\gamma_h$ and linear readout $\bm P_h$. The shared keys follow the deterministic scoring rule of \cref{sec:adapt}, and the per-head scale $s_h$ is set before calibration.
We first select the gate proxy $\phi_i$ from the candidate formulations by reconstruction error. Ridge regression of $t_i$ initializes the gate parameters, weighted least squares estimates the correction gain, and ridge regression with an identity prior fits the linear readout. Two refinement rounds then alternately update the gate and readout: we obtain unconstrained gate targets from a small ridge regression on the projected output and clip them to $[\varepsilon,1-\varepsilon]$ ($\varepsilon=0.005$) before the logit transform, refit the gate parameters, and update the readout by ridge regression. Each calibration step solves a ridge or weighted least-squares subproblem, and refinement updates are retained when they reduce the layerwise objective \cref{eq:relmse} (\cref{alg:et}).

\paragraph{Error decomposition.}
Combining the exact attention decomposition \cref{eq:decomp} with the hybrid attention output \cref{eq:hybrid} yields the error identity
\begin{equation}
\boxed{\bm y_i-\hat{\bm y}_i=(1-m_i)\big(\bm y_i^{\mathcal R}-\hat{\bm y}_i^{\mathcal R}\big)
+(m_i-g_i)\big(\bm y_i^{\mathcal E}-\hat{\bm y}_i^{\mathcal R}\big),}
\label{eq:errdecomp}
\end{equation}
This decomposition separates the total attention error into remote-output reconstruction error and gate estimation error. These terms motivate the complementary roles of the exact branch, mean anchoring, global correction and analytical calibration.

\paragraph{Optional LoRA recovery.}
As errors may accumulate across layers and sampling steps in generative models, an optional stage trains LoRA adapters together with the new parameters, starting from the calibrated model; it is separate from the gradient-free conversion and is the only stage that uses gradients.
\section{Experiments}
\label{sec:exp}\label{sec:main_results}\label{sec:dev}

\paragraph{Experimental setup.}
\label{sec:setup}
We evaluate \ours{} on image classification, class-conditional generation and text-to-image generation for conversion fidelity, optimization cost and inference efficiency. Unless otherwise specified, we use NAT7+32 exact attention, the exact remote value mean, scan-free Global correction, mass-initialized gating and two rounds of alternating least squares, denoted \etpG{} (\etpD{}: Delta-rule scan variant). Gradient-free conversion keeps all pretrained weights frozen and calibrates only the new parameters on unlabeled training samples. Optional recovery trains rank-8 LoRA adapters together with the new parameters. Following \tfive{}~\cite{t5paper}, we report computational cost as FLOPs counted in multiply--accumulates (MACs), with one multiply--accumulate counted as one operation.

\subsection{Image Classification}
\label{sec:cls}
\begin{table}[t]
\centering\small
\setlength{\tabcolsep}{3pt}
\fitto{\linewidth}{\begin{tabular}{@{}l|ccccc@{}}
\toprule
Model & Epochs & Ratio & Top-1 & \#Params & FLOPs \\
\midrule
\rowcolor{etPeach}DeiT-T$^\dagger$~\cite{deit} & 300 & 100\% & 72.05 & 5.7M & 1.25G \\
\midrule
Linear~\cite{linear_transformers} & 30 & 10\% & 63.30 & 5.7M & 1.13G \\
LiT~\cite{lit} & 30 & 10\% & 69.52 & 5.7M & 1.14G \\
CLEAR~\cite{clear} & 30 & 10\% & 68.66 & 5.7M & 1.11G \\
Hedgehog~\cite{hedgehog} & 30 & 10\% & NaN & 5.8M & 1.25G \\
\tfive{}-T~\cite{t5paper} & 30 & 10\% & 71.19 & 6.2M & 1.34G \\
\tfivep{}-T~\cite{t5paper} & 30 & 10\% & \best{72.06} & 6.2M & 1.39G \\
\midrule
\tfive{}-T~\cite{t5paper} & 1 & 0.33\% & 65.57 & 6.2M & 1.34G \\
\tfivep{}-T~\cite{t5paper} & 1 & 0.33\% & 69.30 & 6.2M & 1.36G \\
\midrule
MA-T~\cite{monarchattention} & 0 & 0\% & 66.53 & 5.7M & 1.15G \\
FNA-T~\cite{lu2025artifacts} & 0 & 0\% & 69.54 & 5.7M & 1.73G \\
Nystr\"omformer~\cite{nystromformer} & 0 & 0\% & 67.03 & 5.7M & 1.43G \\
\rowcolor{etSky}\etpG{}-T\oursmark{} & 0 & 0\% & \second{71.66} & 5.9M & 1.26G \\
\bottomrule
\end{tabular}}
\caption{Image classification results on ImageNet-1K. DeiT-T$^\dagger$ is the GAP DeiT-T released by \tfive{}~\cite{t5paper}; Epochs and Ratio count gradient-based updates.}
\label{tab:t5cls}\label{tab:deit_tiny_variants}
\end{table}
\begin{table}[t]
\centering\small
\setlength{\tabcolsep}{3pt}
\fitto{\linewidth}{\begin{tabular}{@{}l|ccccc@{}}
\toprule
Model & Epochs & Ratio & Top-1 & \#Params & FLOPs \\
\midrule
\rowcolor{etPeach}DeiT-S~\cite{deit} & 300 & 100\% & 79.83 & 22.1M & 4.60G \\
\midrule
\tfive{}-S~\cite{t5paper} & 1 & 0.33\% & 73.68 & 23.0M & 4.78G \\
\tfivep{}-S~\cite{t5paper} & 1 & 0.33\% & \second{77.26} & 23.0M & 4.82G \\
\midrule
MA-S~\cite{monarchattention} & 0 & 0\% & 73.72 & 22.1M & 4.42G \\
FNA-S~\cite{lu2025artifacts} & 0 & 0\% & 74.67 & 22.1M & 5.55G \\
\rowcolor{etSky}\etpG{}-S\oursmark{} & 0 & 0\% & \best{79.64} & 22.4M & 4.63G \\
\midrule
\rowcolor{etPeach}DeiT-B~\cite{deit} & 300 & 100\% & 81.80 & 86.6M & 17.56G \\
\midrule
\tfive{}-B~\cite{t5paper} & 1 & 0.33\% & 77.30 & 88.5M & 17.93G \\
\tfivep{}-B~\cite{t5paper} & 1 & 0.33\% & \second{79.91} & 88.5M & 18.02G \\
\midrule
MA-B~\cite{monarchattention} & 0 & 0\% & 79.50 & 86.6M & 17.21G \\
FNA-B~\cite{lu2025artifacts} & 0 & 0\% & 79.78 & 86.6M & 19.47G \\
\rowcolor{etSky}\etpG{}-B\oursmark{} & 0 & 0\% & \best{81.55} & 87.3M & 17.64G \\
\bottomrule
\end{tabular}}
\caption{Image classification results on ImageNet-1K with DeiT-S and DeiT-B. Epochs and Ratio count gradient-based updates.}
\label{tab:cls_clsmain}
\end{table}
\begin{table}[t]
\centering\small
\setlength{\tabcolsep}{3pt}
\fitto{\linewidth}{\begin{tabular}{@{}l|ccccc@{}}
\toprule
Teacher & Softmax & MA & FNA & LoLCATs$^\ddagger$ & \oursplain{}\oursmark{} \\
\midrule
DeiT-III-L & 84.78 & 83.29 & 83.65 & \second{83.88} & \best{84.60} \\
DINOv2-L$^v$ & 77.99 & 59.51 & \second{73.48} & 72.95 & \best{77.74} \\
CLIP-L$^z$ & 73.47 & 59.24 & 49.32 & \second{70.26} & \best{72.53} \\
SigLIP-L$^z$ & 78.63 & 74.75 & 58.87 & \second{77.10} & \best{77.78} \\
DeiT-B@$384^2$ & 83.10 & 80.52 & 75.30 & \second{81.13} & \best{82.00} \\
\bottomrule
\end{tabular}}
\caption{Image classification results of larger and non-supervised teachers. $^\ddagger$2,560 attention-transfer and 2,560 LoRA steps; $^v$ImageNetV2; $^z$zero-shot head.}
\label{tab:cls_cross_main}
\end{table}
\paragraph{Setup.}
We first evaluate conversion fidelity at native input resolutions after replacing all softmax attention layers. We convert DeiT-T, DeiT-S and DeiT-B on ImageNet-1K with 1,024 unlabeled training images. Following \tfive{}, we use the global-average-pooling (GAP) DeiT-T released with \tfive{} and the official class-token (CLS) DeiT-S/B, and compare against \tfive{}, \tfivep{}, MonarchAttention (MA)~\cite{monarchattention} and Fast Nystr\"om Attention (FNA)~\cite{lu2025artifacts}. The one-epoch \tfive{}/\tfivep{} controls fine-tune all parameters for 625 updates with the \tfive{} recipe and the best of three learning rates selected on training-split data.

\paragraph{High-fidelity conversion.}
As shown in \cref{tab:t5cls,tab:cls_clsmain}, without gradient updates \ours{} achieves 71.66\%, 79.64\% and 81.55\% top-1 on DeiT-T, DeiT-S and DeiT-B, within 0.39, 0.19 and 0.25 points of the softmax teachers even though every attention layer is converted; across three calibration-image draws, the standard deviation is at most 0.05 points. \ours{} outperforms the data-free baselines and the one-epoch \tfive{}/\tfivep{} controls, improving over \tfivep{} by 2.38 and 1.64 points on DeiT-S and DeiT-B without fine-tuning. On GAP DeiT-T, its 71.66\% lies between the 30-epoch \tfive{} and \tfivep{} results of 71.19\% and 72.06\%. \ours{} adds only 0.15--0.7M parameters, with FLOPs within about 1\% of the teachers at $224{\times}224$.

\paragraph{Transfer across pretrained architectures.}
On DeiT-III-L, DINOv2-L, CLIP-L, SigLIP-L and DeiT-B at $384{\times}384$ (\cref{tab:cls_cross_main}), \ours{} loses only 0.18--1.10 points and outperforms the conversion baselines; on DINOv2-L it retains 77.74\% against the teacher's 77.99\%, whereas MA and FNA reach 59.51\% and 73.48\%. \ours{} also outperforms LoLCATs, which uses 2,560 attention-transfer and 2,560 LoRA updates; on one A100, \ours{} calibrates these models in 3.4--4.3\,min, against 8.7--11.8\,min for the attention-transfer stage of LoLCATs alone.

\subsection{Class-Conditional Image Generation}
\label{sec:gen}
\begin{table}[t]
\centering
\setlength{\tabcolsep}{3pt}
\small
\fitto{\linewidth}{\begin{tabular}{l|rrrrcc}
\toprule
Model & Epochs & FT-Ratio & FID$\downarrow$ & IS$\uparrow$ & Replace & Distill \\
\midrule
\rowcolor{etPeach} DiT-XL/2~\cite{dit} & 1400 & -- & 2.27 & 278.2 & -- & -- \\
\midrule
LiT-XL/2~\cite{lit} & 280 & 20\% & \best{2.32} & 265.2 & 100\% & \ding{51} \\
Hyena-X~\cite{grafting} & $\sim$28$^\dagger$ & 2\% & 2.74 & 273.30 & 50\% & \ding{55} \\
Hyena-Y~\cite{grafting} & $\sim$28$^\dagger$ & 2\% & 2.72 & 273.37 & 50\% & \ding{55} \\
DiT$^5$-XL/2~\cite{t5paper} & 8 & 0.57\% & 2.48 & 267.9 & 100\% & \ding{55} \\
\midrule
\rowcolor{lolGroup}\multicolumn{7}{@{}l}{\textbf{Gradient-free conversion}} \\
MA-XL/2~\cite{monarchattention} & 0 & 0\% & 20.32 & 115.5 & 100\% & \ding{55} \\
FNA-XL/2~\cite{lu2025artifacts} & 0 & 0\% & 378.52 & 2.5 & 100\% & \ding{55} \\
\rowcolor{etSky}\etpG{}-XL/2\oursmark{} & 0 & 0\% & 5.61 & 197.1 & 100\% & \ding{55} \\
\whiterule
\rowcolor{etSky}\etpD{}-XL/2\oursmark{} & 0 & 0\% & 4.82 & 206.1 & 100\% & \ding{55} \\
\midrule
\rowcolor{lolGroup}\multicolumn{7}{@{}l}{\textbf{250 LoRA updates}} \\
\rowcolor{etPeach} DiT-XL/2-LoRA & 0.05 & 0.0036\% & 2.35 & 261.7 & -- & \ding{55} \\
DiT$^5$-XL/2-LoRA (1$\times$) & 0.05 & 0.0036\% & 47.92 & 53.4 & 100\% & \ding{55} \\
DiT$^5$-XL/2-LoRA (10$\times$) & 0.05 & 0.0036\% & 12.78 & 125.8 & 100\% & \ding{55} \\
\rowcolor{etSky}\etpG{}-XL/2-LoRA\oursmark{} & 0.05 & 0.0036\% & 2.59 & 281.8 & 100\% & \ding{55} \\
\whiterule
\rowcolor{etSky}\etpD{}-XL/2-LoRA\oursmark{} & 0.05 & 0.0036\% & 2.59 & 282.6 & 100\% & \ding{55} \\
\midrule
\rowcolor{lolGroup}\multicolumn{7}{@{}l}{\textbf{1,000 LoRA updates}} \\
\rowcolor{etPeach} DiT-XL/2-LoRA & 0.2 & 0.014\% & 2.26 & 262.9 & -- & \ding{55} \\
DiT$^5$-XL/2-LoRA (1$\times$) & 0.2 & 0.014\% & 15.94 & 115.0 & 100\% & \ding{55} \\
DiT$^5$-XL/2-LoRA (10$\times$) & 0.2 & 0.014\% & 7.66 & 167.5 & 100\% & \ding{55} \\
\rowcolor{etSky}\etpG{}-XL/2-LoRA\oursmark{} & 0.2 & 0.014\% & \second{2.36} & 269.8 & 100\% & \ding{55} \\
\whiterule
\rowcolor{etSky}\etpD{}-XL/2-LoRA\oursmark{} & 0.2 & 0.014\% & \second{2.36} & 268.2 & 100\% & \ding{55} \\
\bottomrule
\end{tabular}}
\caption{Comparison on class-conditional ImageNet $256{\times}256$. DiT-XL/2-LoRA is the softmax teacher trained with the same rank-8 LoRA recipe; for DiT$^5$-LoRA, 1$\times$/10$\times$ is the learning-rate multiplier of its new TTT parameters. FT: fine-tuning. $^\dagger$Two-stage training.}
\label{tab:t5dit}
\end{table}
\begin{table*}[t]
\centering\small
\newcommand{\inh}[1]{\textcolor{etGray}{#1}}
\setlength{\tabcolsep}{3.4pt}
\renewcommand{\arraystretch}{1.08}
\fitto{\textwidth}{\begin{tabular}{@{}l| cccccc ccccccc cc@{}}
\toprule
& \multicolumn{6}{c}{DPG-Bench$\uparrow$} & \multicolumn{7}{c}{GenEval$\uparrow$} & \multicolumn{2}{c}{Latency (s)$\downarrow$} \\
\cmidrule(lr){2-7}\cmidrule(lr){8-14}\cmidrule(l){15-16}
Method & Global & Entity & Attribute & Relation & Other & Overall & Single & Two & Count & Color & Position & Color Attri. & Overall & 1024 & 2048 \\
\midrule
\rowcolor{etPeach} SD3.5-Medium~\cite{sd3} & 84.42 & 89.92 & 88.35 & 93.29 & 80.20 & 84.16 & 1.00 & 0.84 & 0.60 & 0.83 & 0.20 & 0.53 & 0.67 & 25.70 & 257.26 \\
SD3.5-MA~\cite{monarchattention} & 81.46 & 85.74 & 84.16 & 91.72 & 78.50 & 78.37 & \second{0.95} & 0.64 & 0.47 & \second{0.75} & \second{0.15} & 0.43 & 0.57 & -- & -- \\
\rowcolor{etSky} SD3.5-\etpG{}\oursmark{} & \best{85.11} & \second{89.85} & \second{87.88} & \second{93.33} & \second{79.90} & \second{84.04} & \best{1.00} & \second{0.82} & \second{0.62} & \best{0.81} & \best{0.21} & \second{0.54} & \second{0.67} & \best{18.55} & \best{153.40} \\
\whiterule
\rowcolor{etSky} SD3.5-\etpD{}\oursmark{} & \second{84.35} & \best{90.09} & \best{88.25} & \best{93.91} & \best{80.90} & \best{84.52} & \best{1.00} & \best{0.83} & \best{0.67} & \best{0.81} & \best{0.21} & \best{0.55} & \best{0.68} & \second{19.33} & \second{156.66} \\
$\Delta$ (\etpG{} vs.\ Medium) & $+$0.69 & $-$0.07 & $-$0.47 & $+$0.04 & $-$0.30 & $-$0.12 & 0.00 & $-$0.02 & $+$0.02 & $-$0.02 & $+$0.01 & $+$0.01 & 0.00 & 1.39$\times$ & 1.68$\times$ \\
\midrule
\rowcolor{etPeach} SD3.5-Medium-LoRA & 84.35 & 90.77 & 88.67 & 93.41 & 83.60 & 85.08 & 0.99 & 0.89 & 0.72 & 0.85 & 0.23 & 0.62 & 0.72 & \inh{25.70} & \inh{257.26} \\
SD3.5-\tfive{}-LoRA~\cite{t5paper} & 81.69 & \best{90.68} & 87.49 & \second{93.75} & \best{86.30} & 84.56 & \best{1.00} & \second{0.88} & 0.67 & \second{0.82} & 0.23 & \second{0.49} & 0.68 & \inh{19.45} & \inh{156.73} \\
\rowcolor{etSky} SD3.5-\etpG{}-LoRA\oursmark{} & \best{84.42} & \best{90.68} & \best{88.76} & \best{93.82} & 83.30 & \second{85.07} & \best{1.00} & \best{0.89} & \second{0.70} & \best{0.84} & \second{0.25} & \best{0.62} & \second{0.71} & \inh{18.55} & \inh{153.40} \\
\whiterule
\rowcolor{etSky} SD3.5-\etpD{}-LoRA\oursmark{} & \second{84.19} & \second{90.67} & \second{88.70} & 93.66 & \second{83.50} & \best{85.08} & \best{1.00} & \second{0.88} & \best{0.73} & \best{0.84} & \best{0.26} & \best{0.62} & \best{0.72} & \inh{19.33} & \inh{156.66} \\
\bottomrule
\end{tabular}}
\vspace{-0.15cm}
\caption{Quantitative comparison between SD3.5-Medium and its converted models on DPG-Bench and GenEval. We also report latency (seconds) at different resolutions. -LoRA rows use 800 rank-8 LoRA updates; SD3.5-Medium-LoRA is the teacher trained with the same recipe.}
\label{tab:t5sd35}
\vspace{-0.2cm}
\end{table*}
\begin{figure*}[t]
\centering
\includegraphics[width=\textwidth]{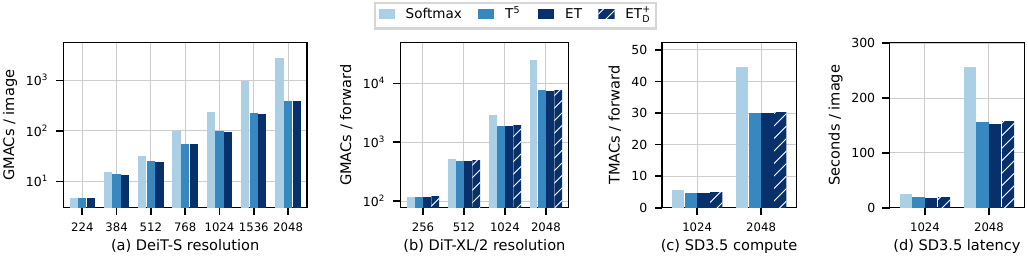}
\vspace{-0.5cm}
\caption{MACs and latency versus resolution on DeiT-S (a), DiT-XL/2 (b) and SD3.5-Medium (c, d). Latency is measured on one H200 GPU. The efficiency advantage of \ours{} becomes more pronounced as the sequence length grows.}
\label{fig:eff_combined}\label{fig:flops_t5_main}\label{fig:effmain}
\vspace{-0.2cm}
\end{figure*}
\paragraph{Setup.}
We convert all 28 attention layers of the pretrained DiT-XL/2~\cite{dit} for class-conditional ImageNet $256{\times}256$ generation. Calibration uses 384 unlabeled training images and four timesteps, layer by layer on the partially converted model to mitigate error accumulation. Following DiT$^5$~\cite{t5paper}, we generate 50K samples with 250 denoising diffusion probabilistic model (DDPM) steps and guidance scale 1.5 and report FID-50K and Inception Score (IS). For optional recovery, rank-8 LoRA starts from the calibrated model and runs 250 or 1,000 updates without exponential moving average (EMA) weights.

\paragraph{Gradient-free conversion.}
As shown in \cref{tab:t5dit}, \etpG{} achieves FID-50K 5.61 without gradient updates, far better than MA (20.32) and FNA (378.52), which also replace all attention layers. \etpD{} further reduces FID-50K to 4.82 without gradient updates, providing an effective initialization for lightweight LoRA recovery.

\paragraph{Low-cost performance recovery.}
With 250 recovery updates using rank-8 LoRA and the newly introduced \ours{} parameters, FID-50K drops from 5.61 to 2.59, compared with 2.35 for the softmax teacher under the same update count and LoRA rank; 1,000 updates reach 2.36, against 2.26 for the teacher (both \etpG{} and \etpD{} reach 2.59 and 2.36). With the same update counts, DiT$^5$ reaches 12.78 and 7.66 with a 10$\times$ learning rate on its new parameters. A high-fidelity analytical initialization thus recovers performance through lightweight adaptation.

\begin{table}[t]
\centering\small
\setlength{\tabcolsep}{3pt}
\fitto{\linewidth}{\begin{tabular}{@{}ll|rrr@{}}
\toprule
Model & Method & Updates & A100$\cdot$h & Quality \\
\midrule
DeiT-S & \tfive{} / \tfivep{} & 625 & 1.08 / 2.09 & 73.68 / 77.26 \\
\rowcolor{etSky} & \etpG{}\oursmark{} & 0 & \best{0.009} & \best{79.64} \\
\midrule
DeiT-B & \tfive{} / \tfivep{} & 625 & 3.09 / 5.48 & 77.30 / 79.91 \\
\rowcolor{etSky} & \etpG{}\oursmark{} & 0 & \best{0.019} & \best{81.55} \\
\midrule
DiT-XL/2 & DiT$^5$ & 8 ep & -- & 2.48 \\
 & DiT$^5$-LoRA (1$\times$ / 10$\times$) & 1000 & 1.71 & 15.94 / 7.66 \\
\rowcolor{etSky} & \etpG{}-LoRA\oursmark{} & 250 & \best{0.45} & 2.59 \\
\rowcolor{etSky} & \etpG{}-LoRA\oursmark{} & 1000 & 1.42 & \best{2.36} \\
\midrule
SD3.5-M & \tfive{}-LoRA & 800 & 8.9 & 0.68 \\
\rowcolor{etSky} & \etpG{}\oursmark{} & 0 & \best{0.73} & 0.67 \\
\rowcolor{etSky} & \etpG{}-LoRA\oursmark{} & 800 & 10.3 & \best{0.71} \\
\bottomrule
\end{tabular}}
\caption{Conversion and recovery costs on one A100-80GB GPU. A100 GPU-hours include gradient-free calibration and any optional LoRA recovery. \ours{} performs data-dependent calibration without gradient updates, whereas \tfive{}/\tfivep{} use gradient-based adaptation. Quality is measured by ImageNet top-1 accuracy, FID-50K, or GenEval, depending on the task.}
\label{tab:conv_cost}
\end{table}
\begin{table}[t]
\centering
\setlength{\tabcolsep}{4pt}
\small
\fitto{\linewidth}{\begin{tabular}{l|ccc}
\toprule
& DeiT-T & DeiT-S & DeiT-B@$384^2$ \\
\midrule
\rowcolor{etPeach}Softmax teacher & 72.13 & 79.83 & 83.10 \\
\midrule
\multicolumn{4}{l}{\textit{(a) Remote branch} (NAT7+32 exact set)} \\
Exact set only (no remote branch) & 67.69 & 78.16 & 78.92 \\
Remote mean only & \second{70.54} & \second{78.99} & 81.05 \\
Global correction only (no mean) & 70.35 & 78.88 & \second{81.45} \\
\rowcolor{etSky}Mean + Global correction\oursmark{} & \best{71.76} & \best{79.64} & \best{82.00} \\
\midrule
\multicolumn{4}{l}{\textit{(b) Calibration of the new parameters}} \\
No fitting (default values) & 66.98 & 77.37 & 80.28 \\
Least squares, no refinement & \second{71.74} & \second{79.63} & \second{81.69} \\
\rowcolor{etSky}+ two refinement rounds\oursmark{} & \best{71.76} & \best{79.64} & \best{82.00} \\
\midrule
\multicolumn{4}{l}{\textit{(c) Exact set} \quad (Global / Delta-rule scan)} \\
\rowcolor{etSky}NAT7+32\oursmark{} & 71.76 / 71.77 & 79.64 / 79.63 & 82.00 / 82.03 \\
NAT5 window only & 70.51 / 70.56 & 78.44 / 78.59 & 78.78 / 79.23 \\
32 shared keys only & 54.10 / 63.69 & 65.70 / 73.62 & 60.78 / 73.12 \\
\bottomrule
\end{tabular}}
\caption{Ablation of \ours{} components on ImageNet-1K. All \ours{} variants use gradient-free conversion with no gradient updates. (a) Contributions of remote mean anchoring and global correction. (b) Effect of analytical calibration and refinement. (c) Effect of exact-set coverage and Global versus Delta-rule correction. The DeiT-T results use the CLS checkpoint, unlike the GAP checkpoint in \cref{tab:t5cls}.}
\label{tab:findings_main}\label{tab:ablation}
\end{table}
\begin{table}[t]
\centering\footnotesize
\setlength{\tabcolsep}{2.6pt}
\fitto{\linewidth}{\begin{tabular}{@{}l*{8}{>{\centering\arraybackslash}p{3.0em}}@{}}
\toprule
Unlabeled images $N$ & 1 & 2 & 4 & 8 & 16 & 32 & 128 & 1024 \\
\midrule
\rowcolor{etSky}\etpG{}\oursmark{} top-1 & 56.84 & 72.35 & 77.50 & 78.89 & 79.30 & 79.45 & \second{79.59} & \best{79.64} \\
Gap to DeiT-S & $-$22.99 & $-$7.48 & $-$2.33 & $-$0.94 & $-$0.53 & $-$0.38 & $-$0.24 & $-$0.19 \\
\midrule
\multicolumn{9}{@{}l@{}}{\footnotesize DeiT-S 79.83 \quad Zero-data, all layers: MA-S~\cite{monarchattention} 73.72, FNA-S~\cite{lu2025artifacts} 74.67} \\
\bottomrule
\end{tabular}}
\vspace{-0.15cm}
\caption{Ablation on the number of unlabeled calibration images on DeiT-S.}
\label{tab:calib_images}
\vspace{-0.2cm}
\end{table}
\begin{figure}[t]
\centering
\includegraphics[width=\linewidth]{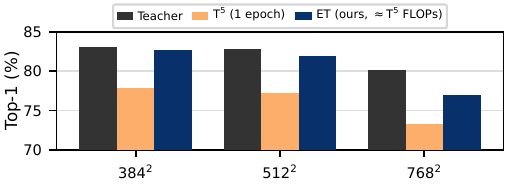}
\vspace{-0.25cm}
\caption{Accuracy--efficiency trade-off across resolutions on DeiT-B. \ours{} uses 191 exact keys with inference FLOPs comparable to \tfive{}. At $768{\times}768$, \ours{} achieves 3.59 percentage points higher top-1 accuracy than \tfive{} while requiring 29\% fewer MACs than the softmax teacher.}
\label{fig:budget}
\vspace{-0.2cm}
\end{figure}
\subsection{Text-to-Image Generation}
\label{sec:sd35}
\paragraph{Setup.}
On Stable Diffusion 3.5 Medium (SD3.5-Medium)~\cite{sd3}, following \tfive{}, we convert the same 18 image-attention modules and keep attention from text queries exact. Calibration uses 32 FLUX-generated images with captions, four timesteps and both conditioning branches. We evaluate GenEval~\cite{geneval} and DPG-Bench~\cite{ella} with 28 sampling steps, guidance scale 3.5 and four images per prompt, and measure latency on one H200 GPU at batch size 1. Optional recovery uses 800 rank-8 LoRA updates on the same data, merged for inference.

\paragraph{Generation quality and inference efficiency.}
As shown in \cref{tab:t5sd35}, gradient-free \etpG{} achieves GenEval 0.666, matching the teacher's 0.665 (paired 95\% interval of the difference $[-0.014,+0.015]$), and DPG-Bench 84.04 against 84.16, while the data-free MA reaches 78.37. This fidelity comes with inference acceleration: latency drops from 25.70\,s to 18.55\,s per image at 1K ($1.39\times$) and from 257.26\,s to 153.40\,s at 2K ($1.68\times$).

\paragraph{Optional LoRA recovery.}
With 800 LoRA updates, \etpG{} reaches GenEval 0.71 and DPG-Bench 85.07, close to the 0.72 and 85.08 of the teacher with the same recipe, which \etpD{} matches; \tfive{}-LoRA reaches 0.68 and 84.56. Merged LoRA keeps the converted architecture and its latency.

\subsection{Efficiency}
\label{sec:efficiency}
\paragraph{Inference efficiency.}
For fixed attention dimensions and exact-set size, \ours{} is linear in sequence length, so its advantage grows with resolution (\cref{fig:eff_combined}): at $2048{\times}2048$, the softmax DeiT-S and DiT-XL/2 need about 7.3$\times$ and 3.3$\times$ the FLOPs of \ours{}, which stays comparable to \tfive{}; on SD3.5-Medium this yields the speedups above.

\paragraph{Conversion efficiency.}
\label{sec:cost}
\ours{} also reduces the optimization cost of conversion (\cref{tab:conv_cost}). On one A100-80GB GPU, calibrating DeiT-S and DeiT-B takes only 31 and 68\,s, against 1.08--2.09 and 3.09--5.48 GPU-hours for the one-epoch \tfive{}/\tfivep{} controls, with higher top-1. On DiT-XL/2, \ours{} reaches FID-50K 2.59 with 250 LoRA updates in 0.45 A100 GPU-hours and 2.36 with 1,000 updates in 1.42\,h. On SD3.5-Medium, gradient-free conversion takes only 0.73 A100 GPU-hours for teacher-level GenEval at 1K, whereas \tfive{}-LoRA needs 8.9\,h to reach 0.68. With 800 further LoRA updates, \ours{} costs 10.3\,h, as its LoRA updates are slightly slower than those of \tfive{}.

\subsection{Ablation Study}
\label{sec:analysis}\label{sec:ablations}\label{sec:find1}\label{sec:recurrence}
\paragraph{Effect of hybrid attention components.}
In \cref{tab:ablation}a, the renormalized exact set alone reaches 78.16\% on DeiT-S; adding only the remote mean or only the global correction gives 78.99\% or 78.88\%, and both give 79.64\%. On DeiT-B at $384{\times}384$, the full remote branch raises 78.92\% to 82.00\%, supporting the complementary roles of mean anchoring and the calibrated correction.

\paragraph{Effect of analytical calibration.}
In \cref{tab:ablation}b, DeiT-S reaches 77.37\% without calibration, 79.63\% with least squares and 79.64\% after two refinement rounds; on DeiT-B at $384{\times}384$, calibration raises 80.28\% to 82.00\%. Adam on the same operator gives comparable accuracy (79.66\% vs.\ 79.64\%) with about 90$\times$ the least-squares fitting time (46\,min).

\paragraph{Exact-set coverage and correction variants.}
In \cref{tab:ablation}c, with NAT7+32, \etpG{} and \etpD{} differ by at most 0.03 points. With only 32 shared keys and no local window, both degrade, \etpG{} more: on DeiT-S, Global falls to 65.70\% and Delta to 73.62\%, and on SD3.5 GenEval drops to 0.03 and 0.40. The scan-free correction is most effective with sufficient exact local interactions, while the Delta-rule scan is more robust when exact coverage is limited.

\paragraph{Calibration data efficiency.}
On DeiT-S (\cref{tab:calib_images}), \ours{} reaches 77.50\% with four unlabeled images, 79.30\% with 16, 79.45\% with 32 (0.38 points below the teacher) and 79.64\% with 1,024.

\paragraph{Accuracy--efficiency trade-off across resolutions.}
Finally, \cref{fig:budget} compares \ours{} with \tfive{} across resolutions, with 191 exact keys for \ours{} on DeiT-B. At $768{\times}768$, \ours{} outperforms \tfive{} by \textbf{3.59 percentage points} at comparable inference FLOPs, while requiring \textbf{29\% fewer MACs than the softmax teacher}; at $384{\times}384$ and $512{\times}512$, it reduces the drops relative to softmax from 5.21 and 5.56 points to 0.36 and 0.89. The advantage persists across the evaluated resolutions, demonstrating a favorable accuracy--efficiency trade-off.
\section{Conclusion}
We presented \ours{}, a framework for transferring pretrained softmax Transformers to efficient hybrid attention architectures with minimal performance degradation and optimization cost. By combining exact local and shared-key attention with mean-anchored global correction and adaptive gating, \ours{} achieves linear complexity in sequence length while preserving important attention interactions. The exact remote value mean provides a stable reference for approximating query-dependent deviations, while scan-free global correction enables efficient information aggregation across distant tokens. To retain pretrained knowledge, \ours{} freezes pretrained weights and calibrates only newly introduced parameters through gradient-free alternating least squares on a small set of unlabeled samples, avoiding costly backpropagation during conversion. Extensive experiments across image classification and generation tasks demonstrate that \ours{} achieves high-fidelity attention conversion with limited calibration data, provides effective initialization for lightweight LoRA recovery, and delivers favorable accuracy–efficiency trade-offs with substantial inference speedups at high resolutions.
\FloatBarrier
{\small
\FloatBarrier
\bibliographystyle{ieeenat_fullname}
\bibliography{references}
}
\end{document}